\pdfoutput=1

\documentclass[11pt]{article}

\usepackage{booktabs}
\usepackage{multirow}
\usepackage{siunitx}
\usepackage{wrapfig}
\usepackage{booktabs}

\usepackage{arxiv}
\usepackage[numbers]{natbib}
\usepackage[table]{xcolor}
\usepackage{times}
\usepackage{latexsym}
\usepackage{wrapfig}
\usepackage[T1]{fontenc}

\usepackage[utf8]{inputenc}

\usepackage{microtype}
\usepackage{graphicx}
\usepackage{subfigure}
\usepackage{booktabs} 

\usepackage{tabularx}  
\usepackage{array} 
\usepackage{bm}
\usepackage{enumitem}
\usepackage[normalem]{ulem}

\usepackage{caption}
\usepackage{algorithmicx}
\usepackage{algpseudocode}
\newcommand{\cmark}{\ensuremath{\checkmark}}
\newcommand{\xmark}{\ensuremath{\times}}

\usepackage[normalem]{ulem}

\algrenewcommand\algorithmicrequire{\textbf{Input:}}
\algrenewcommand\algorithmicensure{\textbf{Output:}}

\usepackage{hyperref}

\usepackage{amsmath}
\usepackage{amssymb}
\usepackage{mathtools}
\usepackage{amsthm}
\usepackage{algorithm}
\usepackage{algpseudocode}

\usepackage{multirow}
\usepackage{multicol}
\usepackage{upgreek}
\usepackage[capitalize,noabbrev]{cleveref}
\theoremstyle{plain}
\newtheorem{theorem}{Theorem}[section]
\newtheorem{proposition}[theorem]{Proposition}

\theoremstyle{definition}

\theoremstyle{remark}

\usepackage[textsize=tiny]{todonotes}

\definecolor{binhgreen}{RGB}{0, 180, 0}

\usepackage[most]{tcolorbox}
\newtcolorbox{graydefbox}{
  enhanced,
  boxrule=0.6pt,
  colback=gray!8,
  colframe=gray!60,
  arc=2mm,
  left=0mm,right=1mm,top=1.5mm,bottom=1.5mm
}

\usepackage{xcolor}
\definecolor{scbg}{gray}{0.88}
\newcommand{\scbox}[1]{\colorbox{scbg}{$\displaystyle #1$}}

\newcommand{\bx}{\bm{x}}

\newcommand{\bM}{{\tt \bm{M}}}

\usepackage[most]{tcolorbox}
\usepackage{paracol}
\usepackage{setspace}

\newtcolorbox{samplebox}[1][]{
  enhanced,
  colback=gray!4,
  colframe=gray!50,
  boxrule=0.35pt,
  arc=1.5pt,
  left=3pt,
  right=3pt,
  top=3pt,
  bottom=3pt,
  fontupper=\ttfamily\footnotesize,
  title=#1,
  fonttitle=\bfseries\footnotesize,
  coltitle=black
}

\usepackage{listings}
\usepackage{xcolor}

\lstdefinestyle{promptstyle}{
  basicstyle=\ttfamily\footnotesize,
  breaklines=true,
  columns=fullflexible,
  frame=single,
  framerule=0.3pt,
  rulecolor=\color{black!30},
  backgroundcolor=\color{gray!5},
  showstringspaces=false,
  keepspaces=true,
  xleftmargin=4pt,
  xrightmargin=4pt,
  aboveskip=6pt,
  belowskip=6pt
}

\allowdisplaybreaks

\title{Simple Self-Conditioning Adaptation for Masked Diffusion Models}

\author{
Michael Cardei\thanks{Equal contribution.} \\
University of Virginia \\
\texttt{ntr2rm@virginia.edu}
\And
Huu Binh Ta\footnotemark[1] \\
University of Virginia \\
\texttt{nzj6jt@virginia.edu}
\And
Ferdinando Fioretto\thanks{Contact author.} \\
University of Virginia \\
\texttt{fioretto@virginia.edu}
}

\hypersetup{
pdftitle={Self-Conditioned Post-Training for Masked Diffusion Models},
pdfsubject={cs.CL, cs.LG},
pdfkeywords={Discrete Diffusion, Generative AI},
}

\begin{document}
\maketitle

\begin{abstract}
Masked diffusion models (MDMs) generate discrete sequences by iterative denoising under an absorbing masking process. In standard MDMs, if a token remains masked after a reverse update, the model discards its clean-state prediction for that position. Thus, still-masked positions must be repeatedly inferred from the mask token alone. This design choice limits cross-step refinement. To address this limitation, this paper introduces \textit{Self-Conditioned Masked Diffusion Models} (SCMDM), a simple and effective \emph{post-training adaptation} that conditions each denoising step on the model's previous clean-state predictions. This is an important departure from partial self-conditioning approaches which requires expensive model training from scratch. In particular, the paper shows that partial self-conditioning, including the commonly used 50\% dropout strategy for training self-conditioned models from scratch, is suboptimal in the post-training regime. Instead, once the model's self-generated clean-state estimates become informative, the specialization to refinement is preferable to mixing conditional and unconditional objectives. SCMDM is evaluated across multiple domains, demonstrating consistent improvement over vanilla MDM baselines, achieving nearly a 50\% reduction in generative perplexity on OWT-trained models (42.89 $\rightarrow$ 23.72), alongside strong improvements in discretized image synthesis quality, small molecular generation, and enhanced fidelity in genomic distribution modeling.
\end{abstract}

\section{Introduction}
\label{sec:introduction}

\begin{wrapfigure}[16]{r}{0.445\textwidth}
    \centering
    \vspace{-1.5em}
    \includegraphics[width=0.445\textwidth]{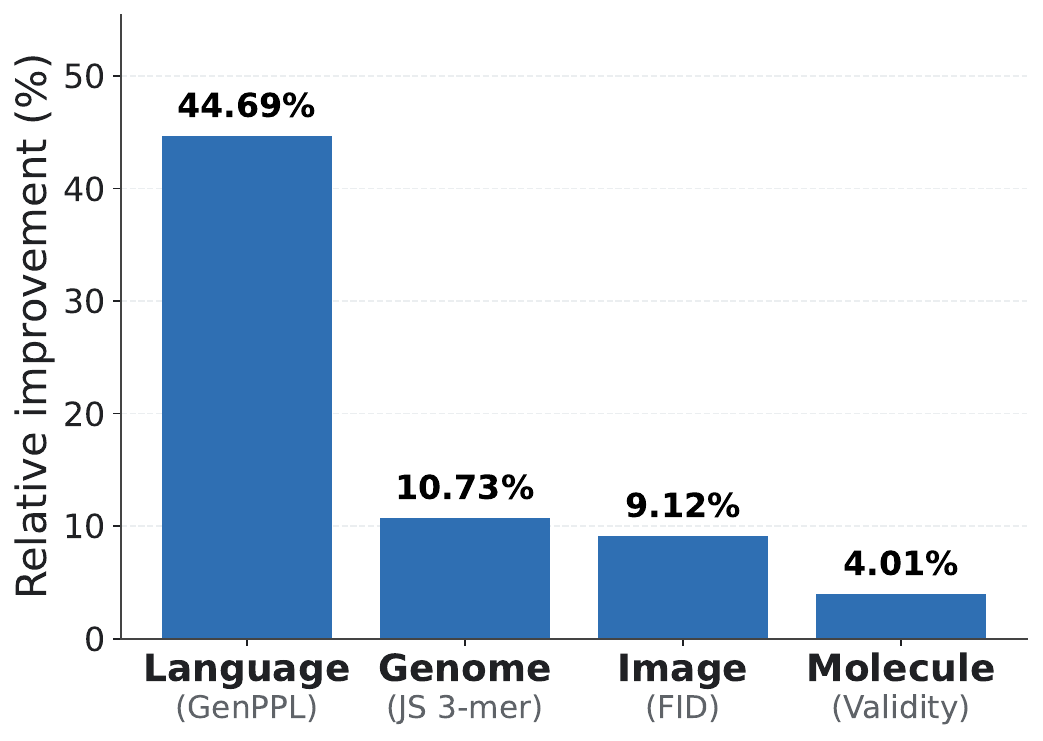}
    \caption{Relative performance boost due to self-conditioning in MDMs.}
    \label{fig:improvement_bar}
\end{wrapfigure}

Masked diffusion models \cite{nie2025largelanguagediffusionmodels,gong2025diffucoderunderstandingimprovingmasked, sahoo2024simpleeffectivemaskeddiffusion} have emerged as a promising framework for discrete sequence generation by replacing left-to-right decoding with iterative denoising over the full sequence. Under an absorbing-mask corruption process, tokens are progressively replaced by a special mask symbol and then reconstructed through repeated denoising updates. 
Beyond competitive modeling performance against autoregressive baselines in natural language and other structured discrete domains \cite{ye2025dream7bdiffusionlarge,bie2025llada20scalingdiffusionlanguage,lee2025genmoldrugdiscoverygeneralist,ta2026search}, MDMs also offer practical advantages. These include parallel token updates which can significantly reduce generation latency \cite{labs2025mercuryultrafastlanguagemodels,christopher2025speculativediffusiondecodingaccelerating,yu2025discretediffusionlargelanguage}, and provide a natural mechanism for iterative refinement and controllable generation \cite{schiff2025simpleguidancemechanismsdiscrete,cardei2025constraineddiscretediffusion}. 

\begin{wrapfigure}[21]{r}{0.45\textwidth}
    \vspace{-12pt}
    \centering
    \includegraphics[width=0.45\textwidth,clip,trim={10 60 10 5}]{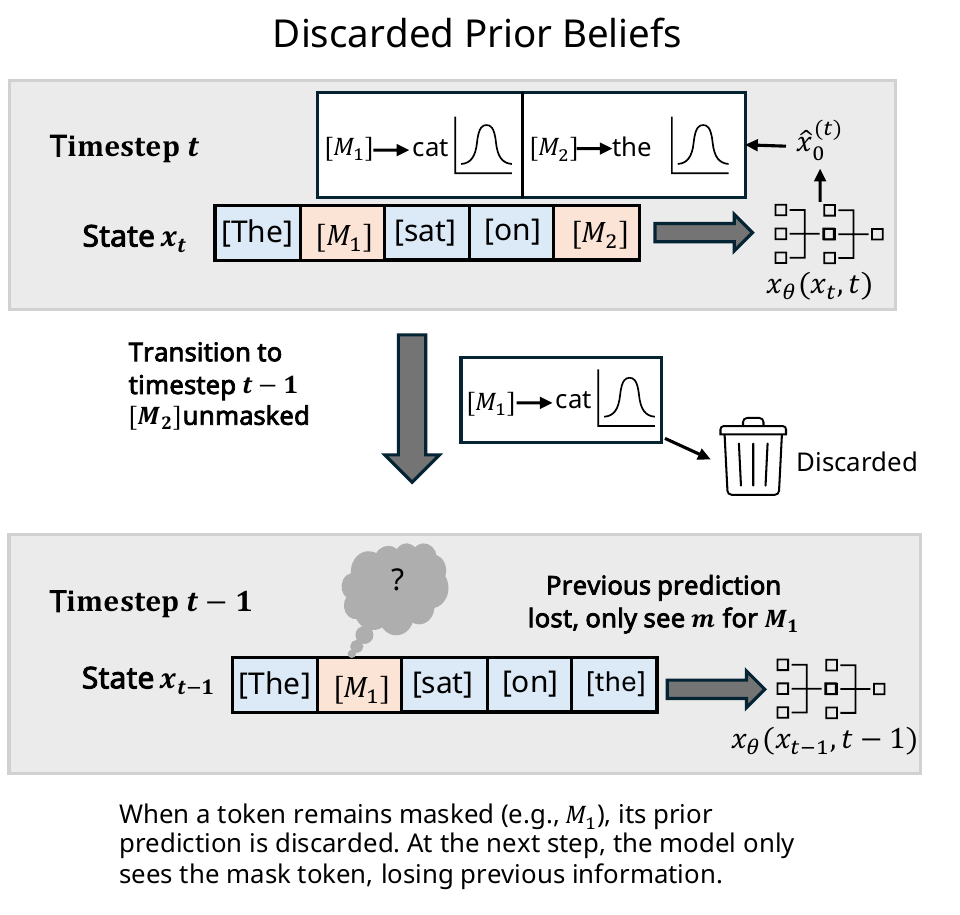}
    \vspace{-12pt}
    \caption{\small Limitation of MDMs: when a token remains masked (e.g., $M_1$) its prior prediction is discarded. At the next step, the model only sees the mask token, losing previous information.}
    \label{fig:issues}
\end{wrapfigure}

However, standard MDM sampling may not fully exploit the model's own predictions across denoising steps \cite{jo2026loopholingdiscretediffusiondeterministic,hu2026residualcontextdiffusionlanguage}. At each reverse step, the model predicts a clean-state distribution for masked positions, but this prediction is discarded after applying the reverse kernel to obtain the next latent state. Thus, if a position remains masked, the model must re-infer it from the mask token alone, rather than refining its earlier estimate across denoising steps (see Figure~\ref{fig:issues}). This limits cross-step reuse of intermediate clean-state predictions, particularly in low-step sampling regimes where each update must make larger progress. 

To address this limitation, we introduce \textit{Self-Conditioned Masked Diffusion Models} (SCMDM), a simple post-training adaptation for absorbing-mask MDMs that conditions each denoising step on the model's previous clean-state predictions. Inspired by self-conditioning techniques in continuous diffusion \cite{strudel2022selfconditionedembeddingdiffusiontext, chen2023analogbitsgeneratingdiscrete}, SCMDM augments the denoiser by feeding its previous clean-state estimate back into the network at each reverse step. 
While recent approaches have started exploring the use of self-conditioning for discrete diffusion models \cite{hu2026residualcontextdiffusionlanguage}, these remain dependent on additional architectural components, require auxiliary reference models, or require training from scratch. In contrast, SCMDM requires only minimal architectural change and can be integrated into existing pretrained MDM backbones as an inexpensive post-training adaptation step, without altering the denoiser evaluations during sampling. 

The core of SCMDM post-training is a two-pass mechanism that trains a single denoiser to first produce an initial clean-state estimate and then refine that estimate under self-conditioning. In the first pass, the model predicts a clean-state distribution from the current masked input alone. In the second pass, this prediction is detached and passed back into the same network as a self-conditioning signal, and the loss is applied only to the resulting refined prediction. This allows SCMDM to adapt a pretrained MDM toward iterative refinement without changing the underlying learning objective.


A key contribution of this work is the identification of a distinct post-training regime for self-conditioning in masked diffusion models. While prior approaches often rely on stochastic unconditional dropout (e.g., a 50/50 mixture of conditioned and unconditioned updates \cite{chen2023analogbitsgeneratingdiscrete}) to stabilize training from scratch, we find that this strategy becomes suboptimal during finetuning of converged backbones. Empirically, we show that once the model's self-generated clean-state estimates are already informative, dropping the self-conditioning signal can be harmful, as it forces training to mix refinement-based updates with less informative unconditional ones. In contrast, finetuning with full self-conditioning consistently yields much stronger adaptation, suggesting this second pass refinement is preferable in the post-training regime.

Empirically, we evaluate SCMDM across natural language generation, molecular generation, genomic sequence generation, and discretized image generation. Across these settings, SCMDM consistently improves over the corresponding masked diffusion baseline. In natural language generation, SCMDM reduces generative perplexity on OpenWebText \cite{Gokaslan2019OpenWeb} from 42.89 to 23.72, providing an impressive 44.69\% improvement. On QM9, it improves molecular generation quality, shown by higher validity (594.2 $\pm$ 9.5 to 618.0 $\pm$ 13.2). In genomic sequence modeling, SCMDM improves distribution fidelity across denoising-step budgets by up to $10.73\%$. For discretized image generation, SCMDM improves FID on CIFAR-10 by 9.12\%. Figure~\ref{fig:improvement_bar} summarizes these gains across domains. Together, these results show that self-conditioning is an effective and broadly applicable post-training adaptation for existing masked diffusion models.

\textbf{Contributions.} This paper makes three key contributions:
    {\bf (1)} It introduces \textit{Self-Conditioned Masked Diffusion Models} (SCMDM), a simple and computationally efficient post-training adaptation for absorbing-mask MDMs that reuses previous clean-state predictions across denoising steps.
    {\bf (2)}  It identifies a distinct post-training regime for self-conditioning: once clean-state estimates become informative, full self-conditioning consistently outperforms partial self-conditioning, revealing a refinement effect and showing that stochastic removal of the self-conditioning signal can be detrimental.
  {\bf (3)}  It demonstrates consistent empirical gains across natural language, molecular, genomic, and discretized image generation tasks.

\newpage
\section{Related Work}

\label{sec:related-work}
\begin{wraptable}[12]{r}{0.56\textwidth}
\vspace{-2.0\baselineskip}
\centering
\footnotesize
\setlength{\tabcolsep}{3pt}
\renewcommand{\arraystretch}{1.05}
\begin{tabularx}{\linewidth}{@{}>{\raggedright\arraybackslash}X c c c@{}}
\toprule
Method
& \shortstack[c]{Absorbing-mask\\ MDM}
& \shortstack[c]{Post-\\training}
& \shortstack[c]{No auxiliary\\ model} \\
\midrule
\textit{Analog Bits} \cite{chen2023analogbitsgeneratingdiscrete}
& \xmark & \xmark & \cmark \\
Embedding diffusion \cite{strudel2022selfconditionedembeddingdiffusiontext}
& \xmark & \xmark & \cmark \\
Loopholing discrete diffusion \cite{jo2026loopholingdiscretediffusiondeterministic}
& \cmark & \xmark & \cmark \\
Residual context diffusion~\cite{hu2026residualcontextdiffusionlanguage}\!\!\!\!\!\!\!\!\!\!\!\!
& \cmark & \xmark & \xmark \\
\rowcolor{gray!15}
SCMDM
& \cmark & \cmark & \cmark \\
\bottomrule
\end{tabularx}
\caption{Comparison with related cross-step information reuse methods. Checkmarks denote properties aligned with lightweight post-training adaptation of pretrained absorbing-mask MDMs.}
\label{tab:related_comparison}
\vspace{-0.8\baselineskip}
\end{wraptable}

\textbf{Discrete Diffusion Models.}
Denoising-based generative modeling has been extended from continuous domains to discrete state spaces through iterative corruption and refinement. Early work by \citet{austin2023structureddenoisingdiffusionmodels} formulated diffusion over discrete variables via structured forward corruption and learned reverse transitions, while \citet{meng2023concretescorematchinggeneralized} and \citet{lou2024discretediffusionmodelingestimating} developed alternative training views based on continuous relaxations and density-ratio estimation. Masked diffusion language models (MDLMs) \cite{sahoo2024simpleeffectivemaskeddiffusion} instantiate this framework with an absorbing-mask corruption process, yielding a scalable non-autoregressive approach to sequence modeling. Recent systems extend this design to large-scale language modeling, code, molecules, and fast generation, including LLaDA \cite{nie2025largelanguagediffusionmodels}, DiffuCoder \cite{gong2025diffucoderunderstandingimprovingmasked}, GenMol \cite{lee2025genmoldrugdiscoverygeneralist}, and Mercury \cite{labs2025mercuryultrafastlanguagemodels}. These works establish masked diffusion as a practical framework for discrete sequence generation. However, standard absorbing-mask MDMs do not retain clean-state predictions for positions that remain masked after a reverse update. At the next denoising step, these positions are again represented only by the \texttt{[MASK]} token, forcing the model to re-infer their clean values rather than refine its previous estimates.

\textbf{Self-Conditioning and Cross-Step Reuse.}
Self-conditioning refers to conditioning a denoiser on its own earlier predictions along the denoising trajectory. For discrete data, \textit{Analog Bits} \cite{chen2023analogbitsgeneratingdiscrete} applies self-conditioning in a continuous bit-diffusion representation, while \citet{strudel2022selfconditionedembeddingdiffusiontext} study self-conditioning for text diffusion in continuous embedding space. These methods show that reusing previous denoising estimates can improve iterative generation, but they do not directly address pretrained absorbing-mask MDMs, where still-masked positions collapse to a single absorbing symbol and the previous clean-state token distribution is discarded.
Recent methods are closer to our setting because they also seek to preserve information across discrete denoising steps. \textit{Loopholing Discrete Diffusion Models} \cite{jo2026loopholingdiscretediffusiondeterministic} introduces a deterministic latent pathway that carries pre-sampling information across steps. However, this mechanism changes the model architecture and is not primarily designed as an efficient post-training retrofit for existing pretrained MDM backbones, instead requires training a new model from scratch with the additional deterministic pathway.
Concurrently, \textit{Residual Context Diffusion Language Models} \cite{hu2026residualcontextdiffusionlanguage}
recycle discarded token representations during diffusion decoding by injecting them back as contextual residuals. This approach also exploits cross-step information reuse, but it requires an additional reference model to construct contextual residuals during training, which increases training complexity and makes the method unsuitable for post-training adaptation.

In contrast, SCMDM is designed as a post-training adaptation of pretrained absorbing-mask MDMs. It introduces no auxiliary model, requires only minimal architectural modification, and reuses the clean-state prediction already computed at step $t+1$ as the self-conditioning input at step $t$, incurring no additional denoiser evaluations at sampling time.
Rather than modifying the diffusion state with a new deterministic pathway or training residual context with a reference model, SCMDM directly conditions the denoiser on its previous clean-state prediction, and enabling masked positions to be refined across denoising steps. We further show that, once the pretrained denoiser is sufficiently informative, full self-conditioning outperforms the partial dropout strategy used in \cite{chen2023analogbitsgeneratingdiscrete}, revealing a refinement-specialization effect not previously characterized. Table \ref{tab:related_comparison} summarizes the key differences between SCMDM and related approaches for cross-step information reuse in discrete diffusion.

\section{Preliminaries}

Let $\bx=(x^1,\dots,x^L)$ denote a length-$L$ sequence over vocabulary $\mathcal V$. For the diffusion process, each token $x^i$ is represented by its one-hot encoding over $\mathcal V$. In natural language generation, $\mathcal{V}$ is a set
of subword tokens (e.g., \texttt{hello}, \texttt{cat}), whereas in biological sequence generation, such as molecular generation, it may be represented as SMILES \cite{weininger1988smiles}, where
$\mathcal{V}$ may consist of chemical symbols and syntax tokens (e.g., \texttt{C}, \texttt{O}). Discrete diffusion models define a Markov noising process which progressively noises an initial sequence $\bm{x}_0$ into a simple prior state $\bm{x}_T$. This time dependent corruption process $\bm{x}_0 \to \bm{x}_T$ is called the forward process and is parameterized by a set of transition probabilities $q(\bm{x}_t \mid \bm{x}_{t-1})$, for $t=1,\ldots,T$. Following the forward noising process, a parametrized reverse process $p_\theta(\bm{x}_{t-1} \mid \bm{x}_t)$ is learned to reverse the corruption $\bm{x}_T \to \bm{x}_0$ and generate samples from the original data distribution.

For masked diffusion models, the forward process is an absorbing corruption mechanism in which tokens are independently replaced with a special $\texttt{[MASK]}$ symbol. The intermediate sequences $\bm{x}_t$ for $t=1,\ldots,T$ are latent variables represented as a sequence of one-hot token states over $\mathcal{V}$; when a position is masked, its token is set to the one-hot vector corresponding to $\texttt{[MASK]}$, which is denoted by ${\tt \bm{M}}$. The forward process from $\bm{x}_0 \to \bm{x}_T$ is represented as:
\begin{equation}
\label{eq:discrete_diffusion_tr_prob}
  q(x_t^i \mid x_{t-1}^i) = \mathrm{Cat}\!\left(x_t^i; \; \alpha_{t\mid t-1} x_{t-1}^i + (1 - \alpha_{t\mid t-1})\bM \right),
\end{equation}
 where $\alpha_{t\mid t-1} := \frac{\alpha_t}{\alpha_{t-1}}$. The noise schedule $\{\alpha_t\}_{t=0}^T$ is monotonically decreasing with $\alpha_0 = 1$ and $\alpha_T = 0$. As shown in Eq.~\eqref{eq:discrete_diffusion_tr_prob}, at each position $i$, at time  $t-1$ unmasked tokens remain unmasked with probability $\alpha_{t\mid t-1}$ and are turned into the \texttt{[MASK]} token with probability $1-\alpha_{t\mid t-1}$. Notably, for the noising process, once a token becomes masked it stays unchanged for the remainder of the forward process. This absorbing forward process admits a closed-form posterior for stepping from timestep $t$ to an earlier timestep $t-1$. However, for the denoising process during generation, $\bm{x}_0$ is not known, as a result, the denoising reverse process is parameterized by a neural network denoiser  $x_\theta(\bm{x}_t,t)$ which is trained to predict a distribution over clean-state tokens $\hat{\bx}_0^{(t)}$, thereby approximating $\bm{x}_0$. The reverse transitions are then represented as:
\begin{equation}
\label{eq:reverse}
p_\theta(x_{t-1}^i \mid \bx_t) =
\begin{cases}
\mathrm{Cat}(x_{t-1}^i; x_t^i), & x_t^i \neq \bM,\\[4pt]
\mathrm{Cat}\!\left(x_{t-1}^i;\dfrac{(1-\alpha_{t-1})\,\bM+(\alpha_{t-1}-\alpha_t)\,x_\theta^i(\bx_t,t)}{1-\alpha_t}\right), & x_t^i = \bM~.
\end{cases}
\end{equation}

across all timesteps and positions $i\in[L]$. 
Once the denoiser $x_\theta$ is trained, sampling occurs by first initializing $\bm{x}_T$ to consist of a fully masked sequence and then proceeds by iteratively denoises for each step $T \to 0$. The final generated sequence is the result of the final denoising step when $t=0$.

\section{Self-Conditioned Masked Diffusion Models}

\begin{figure}[t]
    \centering
    \includegraphics[width=0.85\linewidth]{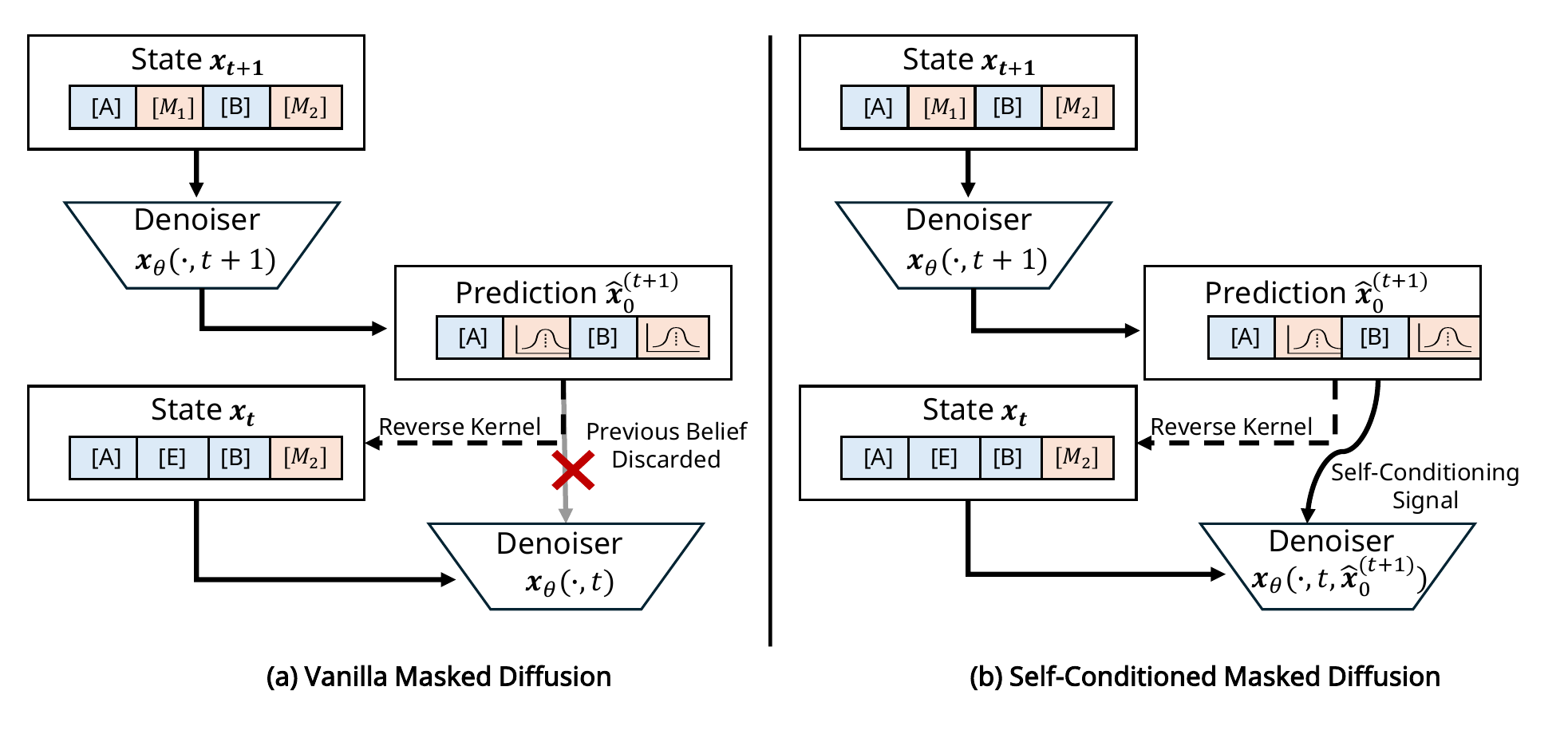}
    \caption{Comparison of reverse denoising steps. (a) Vanilla method discards the clean state prediction after applying the reverse kernel to obtain the next latent state. (b) \emph{SCMDM} conditions the next steps denoiser on both the current latent state and the previous clean state prediction.}
    \label{fig:method}
\end{figure}

The self-conditioned formulation modifies the reverse process so that each denoising step conditions on the current noisy latent state $\bx_t$ together with a clean-state estimate produced at the previous reverse step. During sampling the model's previous clean-state prediction is carried forward
\[
\hat{\bx}_0^{(t+1)} := x_\theta(\bx_{t+1}, t+1),
\]
and used to define a self-conditioned denoiser
\[
\hat{\bx}_{0}^{(t)} := x_\theta(\bx_t, t, \hat{\bx}_0^{(t+1)}).
\]
When no previous clean-state estimate is available, let $\mathbf{0}_{\mathrm{sc}} \in \mathbb{R}^{L \times |\mathcal{V}|}$ denote the null self-conditioning input, i.e., a zero tensor with the same shape as $\hat{\bx}_0^{(t)}$.
This self-conditioned prediction is then substituted into the standard masked diffusion reverse kernel in place of the usual clean-state estimate. Equivalently, the resulting reverse transition at timestep $t$ can be written as
\begin{equation}
\label{eq:scmdm_reverse}
{\small
p_\theta\!\left(x_{t-1}^i \mid \bx_t, \scbox{\hat{\bx}_0^{(t+1)}}\right) =
\begin{cases}
\mathrm{Cat}(x_{t-1}^i; x_t^i), & x_t^i \neq \bM,\\[4pt]
\mathrm{Cat}\!\left(
x_{t-1}^i;
\dfrac{(1-\alpha_{t-1})\,\bM + (\alpha_{t-1}-\alpha_t)\,
x_\theta^i(\bx_t,t,\scbox{\hat{\bx}_0^{(t+1)}})}
{1-\alpha_t}
\right), & x_t^i = \bM~.
\end{cases}
}
\end{equation}
Notably, $\hat{\bx}_0^{(t+1)}$ provides token-level clean-state estimates for positions that would otherwise appear only as $\bM$ in $\bx_t$.

\textbf{Post-training adaptation with a two-pass approximation.}
The setting considered here is post-training adaptation of an existing masked diffusion model. During post-training, timesteps are sampled independently, so the model does not naturally have access to the previous-step estimate $\hat{\bx}_0^{(t+1)}$ that would be available during sampling. Self-conditioning is therefore approximated with two forward passes at the same timestep. First, an intermediate clean-state prediction is computed
\[
\hat{\bx}_0^{(t)} = x_\theta(\bx_t,t, \mathbf{0}_{\mathrm{sc}}),
\]
and gradients are stopped through this estimate, denoted $\mathrm{sg}(\hat{\bx}_0^{(t)})$. This is important because, without stop-gradient, the model could reduce the loss by shaping the first-pass output to assist the second pass, rather than to provide a faithful clean-state estimate. The denoiser is then run a second time, conditioning on the detached prediction,
\begin{equation}
x_\theta(\bx_t,t,\mathrm{sg}(\hat{\bx}_0^{(t)})),
\end{equation}
and this second-pass output is used in the standard masked diffusion training objective. This preserves the original learning objective while adapting the model to exploit an explicit cross-step self-conditioning signal. The standard masked diffusion training objective is retained and applied to this second-pass self-conditioned prediction. Thus, SCMDM changes the denoiser input, but not the underlying learning objective. 
The choice of always conditioning on the detached first-pass prediction rather than stochastically dropping it in favor of 
$\mathbf{0}_{\mathrm{sc}}$, as is common when training self-conditioned diffusion models from  scratch~\cite{chen2023analogbitsgeneratingdiscrete}, reflects a 
structural difference between the post-training and training-from-scratch regimes. 

The complete computational procedures for SCMDM are formalized in Algorithm~\ref{alg:scmdm}. Note that, while the training phase employs a two-pass forward mechanism to internalize the refinement objective, the inference procedure retains an identical amount of denoiser calls to standard MDMs by effectively reusing clean-state predictions from the preceding denoising steps. To maintain a consistent input interface for the denoiser during the two-pass training, the self-conditioning input for the first pass is initialized with the null tensor $\mathbf{0}_{\mathrm{sc}}$. This ensures that the first pass relies solely on the noisy input $\bx_t$ and the pre-trained weights, effectively serving as the model's 'unbiased' initial proposal before the refinement in the second pass.

\textbf{Sampling.}
At inference time, sampling is initialized from the fully masked sequence $\bx_T = (\bM,\ldots,\bM)$ and proceeds from timestep $T$ down to $0$. In this setting, self-conditioning incurs no additional forward passes: after $\hat{\bx}_0^{(t+1)}$ is computed at step $t+1$, it is reused when evaluating the denoiser at step $t$, and the corresponding reverse kernel is then applied to obtain $\bx_{t-1}$. At the first denoising step, no previous clean-state estimate is available, so the self-conditioning input is initialized with the null tensor $\mathbf{0}_{\mathrm{sc}}$. The full sampling process is illustrated in Algorithm \ref{alg:scmdm}.

\textbf{Integrating the self-conditioning signal.}
With the reverse process and the training and sampling procedures specified, the remaining design factor is how the self-conditioning estimate should be integrated into the denoiser. Two ways to inject the self-conditioning estimate into the denoiser are considered. In both cases, the goal is to integrate the representation of the current noisy input $\bx_t$ with a projected version of the clean-state estimate $\hat{\bx}_0^{(t+1)}$ before passing the result through the backbone.

\begin{algorithm}[t]
\caption{SCMDM training and sampling procedures.}
\label{alg:scmdm}
\small
\begin{minipage}[t]{0.475\linewidth}
\textbf{(a) Post-training adaptation}
\vspace{0.25em}
\begin{algorithmic}[1]
\Require pretrained denoiser $x_\theta$, data distribution $q(\bx_0)$, schedule $\{\alpha_t\}_{t=0}^T$
\While{not converged}
    \State sample $\bx_0 \sim q(\bx_0)$ and $t \sim \mathrm{Unif}(\{1,\dots,T\})$
    \State sample $\bx_t \sim q(\bx_t \mid \bx_0)$
    \State $\hat{\bx}_{0,\mathrm{init}}^{(t)} \gets x_\theta(\bx_t,t,\mathbf{0}_{\mathrm{sc}})$
    \State $\tilde{\bx}_0^{(t)} \gets \mathrm{sg}(\hat{\bx}_{0,\mathrm{init}}^{(t)})$
    \State $\hat{\bx}_0^{(t)} \gets x_\theta(\bx_t,t,\tilde{\bx}_0^{(t)})$
    \State compute MDM loss using $\hat{\bx}_0^{(t)}$
    \State update $\theta$
\EndWhile
\end{algorithmic}
\end{minipage}
\hfill
\vrule width 0.4pt
\hfill
\begin{minipage}[t]{0.475\linewidth}
\textbf{(b) Sampling}
\vspace{0.25em}
\begin{algorithmic}[1]
\Require trained denoiser $x_\theta$, schedule $\{\alpha_t\}_{t=0}^T$
\Ensure generated sequence $\bx_0$
\State $\bx_T \gets (\bM,\dots,\bM)$
\State $\tilde{\bx}_0^{(T+1)} \gets \mathbf{0}_{\mathrm{sc}}$
\For{$t=T,T-1,\dots,1$}
    \State $\hat{\bx}_0^{(t)} \gets x_\theta(\bx_t,t,\tilde{\bx}_0^{(t+1)})$
    \State sample $\bx_{t-1}$ from Eq.~\eqref{eq:scmdm_reverse}
    \State $\tilde{\bx}_0^{(t)} \gets \hat{\bx}_0^{(t)}$
\EndFor
\State \Return $\bx_0$
\end{algorithmic}
\end{minipage}
\end{algorithm}

\begin{enumerate}[leftmargin=*, parsep=0.5em, topsep=0.0em]
    \item \textbf{Concatenation-based fusion.} The self-conditioning distribution is
    mapped from vocabulary space to the model hidden space via a linear projection
\(
\mathrm{proj}_{\mathrm{sc}}:\mathbb{R}^{|\mathcal{V}|}\rightarrow\mathbb{R}^{H},
\)
where $|\mathcal{V}|$ is the vocabulary size and $H$ is the hidden dimension. The projected self-conditioning feature is then concatenated with the token embedding of $\bx_t$ and fused through a learned linear layer
\(
\mathrm{fuse}_{\mathrm{proj}}:\mathbb{R}^{2H}\rightarrow\mathbb{R}^{H},
\)
yielding the combined representation passed to the backbone.

\item \textbf{Additive embedding fusion.}
As a lighter alternative, the projected self-conditioning feature can be added directly to the token embedding:
\(
h_t = \mathrm{embed}(\bx_t) + \mathrm{proj}_{\mathrm{sc}}(\hat{\bx}_0^{(t+1)}),
\)
where $\mathrm{embed}(\cdot)$ denotes the token embedding layer. This treats self-conditioning as an auxiliary feature injected into the same latent space as the token embeddings while preserving dimensionality.
\end{enumerate}

To analyze the necessity of full self-conditioning during adaptation, we define $\lambda \in [0, 1]$ as the self-conditioning update ratio, which represents the probability that a training update utilizes the self-conditioning signal. While SCMDM by default employs full self-conditioning ($\lambda = 1$), we introduce this parameter to investigate variants that mix conditional and unconditional updates, allowing us to empirically test whether the stochastic dropout strategy common in training-from-scratch regimes remains beneficial during post-training.

\section{Experiments}
\label{sec:experiments}

\begin{wrapfigure}[15]{r}{0.55\textwidth}
    \vspace{-3em}
    \centering
    \includegraphics[width=0.55\textwidth]{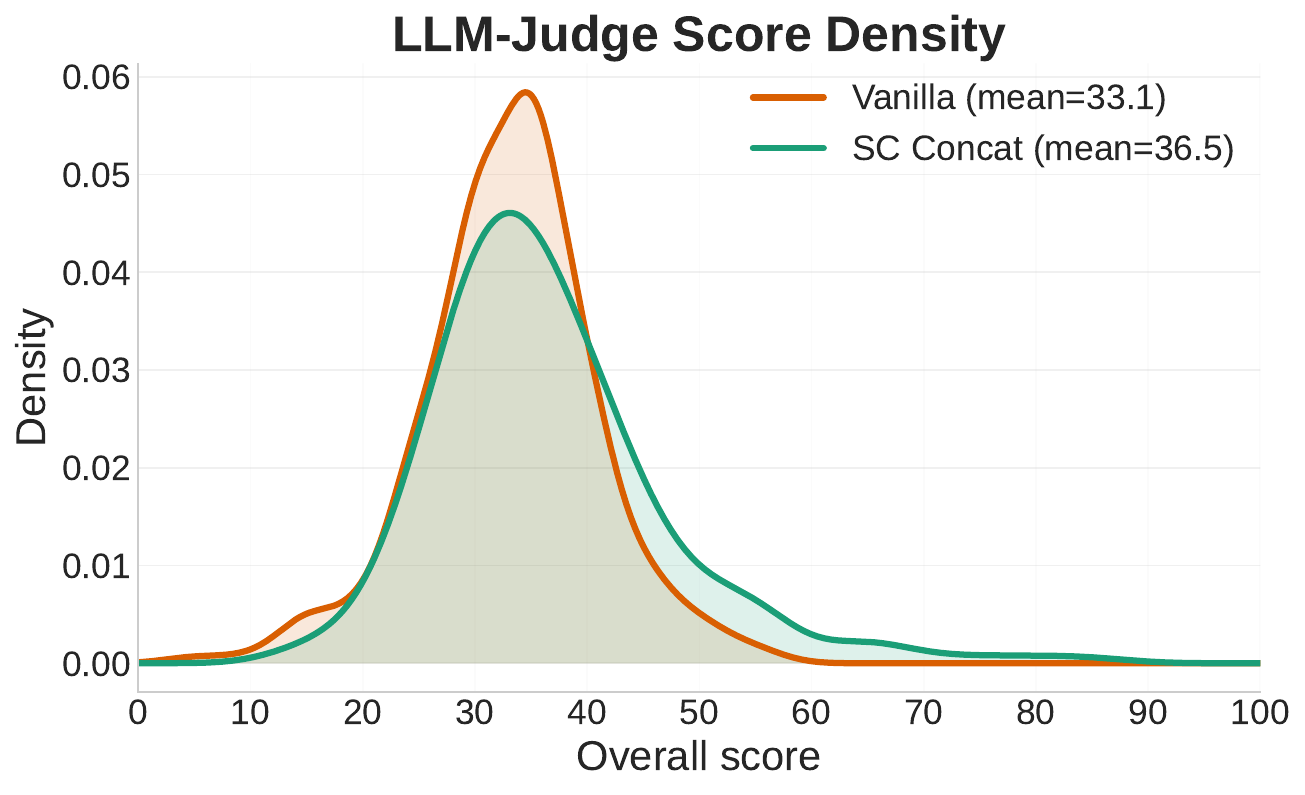}
    \caption{Distribution of LLM-judge scores.}
    \label{fig:llm_judge}
    \vspace{-0.8em}
\end{wrapfigure}

To empirically demonstrate the advantage of SCMDM as a lightweight post-training adaptation, this section evaluates the method across four diverse domains: natural language modeling, molecular generation, genomic sequence generation, and discretized image generation. In each setting, the corresponding vanilla masked diffusion baseline is compared against additive and concatenation-based self-conditioning variants, as well as full versus partial self-conditioning during post-training. These experiments test whether explicit reuse of prior clean-state predictions can reliably improve pretrained absorbing-mask MDMs, and whether the post-training regime favors full self-conditioning over mixed conditional/unconditional training. In addition, we compare SCMDM with LDDMs \cite{jo2026loopholingdiscretediffusiondeterministic} and standard post-training methods using models with at least the same number of parameters as SCMDM (param-match). Notably, across all four domains, SCMDM consistently improves over the vanilla baseline.

\begin{table}[t]
  \centering
  \setlength{\tabcolsep}{5pt}
  \renewcommand{\arraystretch}{1.12}
  \footnotesize
  \begin{tabular}{@{} l
                  S[table-format=4.2]
                  S[table-format=2.2]
                  S[table-format=2.2]
                  S[table-format=2.2]
                  S[table-format=2.2]
                  S[table-format=4.2]
                  S[table-format=2.2]
                  S[table-format=2.2]
                  S[table-format=2.2]
                  S[table-format=2.2] @{}}
    \toprule
    & \multicolumn{5}{c}{\textbf{GPT2-Large evaluator} [$\downarrow$]}
    & \multicolumn{5}{c}{\textbf{Llama3.2-1B evaluator} [$\downarrow$]} \\
    \cmidrule(lr){2-6} \cmidrule(lr){7-11}
    \textbf{Model}
      & \multicolumn{5}{c}{\textbf{Steps}}
      & \multicolumn{5}{c}{\textbf{Steps}} \\
    & {\textbf{1}} & {\textbf{128}} & {\textbf{256}} & {\textbf{512}} & {\textbf{1000}}
    & {\textbf{1}} & {\textbf{128}} & {\textbf{256}} & {\textbf{512}} & {\textbf{1000}} \\
    \midrule
    MDLM
      & 4169.20 & 87.64 & 65.67 & 55.35 & 42.89
      & 2772.38 & 91.32 & 70.49 & 59.73 & 46.90 \\
    $\text{MDLM}_{\text{param-match}}$
      & 4085.08 & 85.60 & 65.70 & 54.92 & 43.96
      & 2729.33 & 89.52 & 69.96 & 59.08 & 48.43 \\
    \midrule
    LDDM-M \cite{jo2026loopholingdiscretediffusiondeterministic}
      & \textbf{3990.01} & 82.86 & 63.65 & 54.83 & 49.01
      & 2664.15 & 85.33 & 64.95 & 54.70 & 49.76 \\
    \midrule
        $\text{SCMDM}_{\text{Concat, }\lambda=0.25}$
      & 4060.36 & 68.57 & 53.38 & 43.97 & 35.49
      & 2639.67 & 71.21 & 56.65 & 46.57 & 38.56 \\
    $\text{SCMDM}_{\text{Concat, }\lambda=0.5}$
      & 4194.39 & 71.79 & 55.84 & 45.98 & 37.04
      & 2666.13 & 74.61 & 59.40 & 48.42 & 40.47 \\
    $\text{SCMDM}_{\text{Concat, }\lambda=0.75}$
      & 4129.75 & 73.49 & 55.31 & 47.15 & 38.37
      & 2668.51 & 75.95 & 58.33 & 49.44 & 41.23 \\
    \midrule
    \rowcolor{gray!15}
    $\text{SCMDM}_{\text{Concat}}$
      & 4059.77 & \textbf{44.59} & \textbf{34.02} & \textbf{27.96} & \textbf{23.72}
      & \textbf{2610.86} & \textbf{45.61} & \textbf{35.54} & 29.16 & \textbf{24.96} \\
    \rowcolor{gray!15}
    $\text{SCMDM}_{\text{Add}}$
      & 4572.87 & 45.47 & 35.22 & 28.15 & 24.48
      & 2986.23 & 46.16 & 36.41 & \textbf{28.90} & 25.48 \\
    \bottomrule
  \end{tabular}
  \caption{Generative perplexity for unconditioned sampling at different denoising steps and under two autoregressive evaluator models (GPT2-Large and Llama3.2-1B).}
  \label{tab:genppl}
\end{table}

\begin{table}[t]
  \centering
  \setlength{\tabcolsep}{3.5pt}
  \renewcommand{\arraystretch}{1.08}
  \footnotesize
  \begin{tabular}{@{} l
                  S[table-format=1.2]
                  S[table-format=1.2]
                  S[table-format=1.2]
                  S[table-format=1.2]
                  S[table-format=1.2]
                  S[table-format=1.3]
                  S[table-format=1.3]
                  S[table-format=1.3]
                  S[table-format=1.3]
                  S[table-format=1.3]
                  S[table-format=1.3]
                  S[table-format=1.3]
                  S[table-format=1.3] @{}}
    \toprule
    & \multicolumn{5}{c}{\textbf{Entropy} [$\uparrow$]}
    & \multicolumn{4}{c}{\textbf{L-AR@1}}
    & \multicolumn{4}{c}{\textbf{G-AR@4}} \\
    \cmidrule(lr){2-6} \cmidrule(lr){7-10} \cmidrule(lr){11-14}
    \textbf{Model}
    & {\textbf{1}} & {\textbf{128}} & {\textbf{256}} & {\textbf{512}} & {\textbf{1000}}
    & {\textbf{128}} & {\textbf{256}} & {\textbf{512}} & {\textbf{1000}}
    & {\textbf{128}} & {\textbf{256}} & {\textbf{512}} & {\textbf{1000}} \\
    \midrule
    MDLM
      & 5.88 & \textbf{5.55} & 5.45 & \textbf{5.39} & 5.28
      & 0.000 & 0.000 & 0.000 & 0.001
      & 0.029 & 0.027 & 0.026 & 0.026 \\
     $\text{MDLM}_{\text{param-match}}$
      & 5.86 & 5.54 & \textbf{5.46} & \textbf{5.39} & \textbf{5.29}
      & 0.000 & 0.000 & 0.000 & 0.001
      & 0.029 & 0.027 & 0.026 & 0.025 \\
    \cmidrule(lr){1-14}
    $\text{SCMDM}_{\text{Concat, }\lambda=0.25}$
      & 5.85 & 5.51 & 5.42 & 5.35 & 5.26
      & 0.000 & 0.000 & 0.000 & 0.001
      & 0.029 & 0.027 & 0.026 & 0.025 \\
    $\text{SCMDM}_{\text{Concat, }\lambda=0.5}$
      & 5.86 & 5.52 & 5.44 & 5.37 & 5.27
      & 0.000 & 0.000 & 0.000 & 0.001
      & 0.029 & 0.027 & 0.026 & 0.025 \\
    $\text{SCMDM}_{\text{Concat, }\lambda=0.75}$
      & 5.86 & 5.53 & 5.43 & 5.37 & 5.28
      & 0.000 & 0.000 & 0.000 & 0.001
      & 0.029 & 0.027 & 0.026 & 0.025 \\
    \cmidrule(lr){1-14}
    \rowcolor{gray!15}
    $\text{SCMDM}_{\text{Concat}}$
      & 5.85 & 5.50 & 5.41 & 5.35 & 5.28
      & 0.000 & 0.000 & 0.000 & 0.001
      & 0.029 & 0.026 & 0.026 & 0.026 \\
    \rowcolor{gray!15}
    $\text{SCMDM}_{\text{Add}}$
      & \textbf{5.90} & 5.50 & 5.43 & 5.35 & \textbf{5.29}
      & 0.000 & 0.000 & 0.000 & 0.001
      & 0.029 & 0.027 & 0.026 & 0.026 \\
    \bottomrule
  \end{tabular}
  \caption{Entropy and autoregressiveness statistics for unconditioned sampling at different denoising budgets. Entropy denotes token-level entropy of the generated sequence. L-AR@1 and G-AR@4 denote local and global autoregressiveness metrics. Higher AR values indicate more autoregressive-like behavior.}
  \label{tab:entropy_ar_pretty}
\end{table}

\subsection{Natural Language}
\label{exp:NL}


SCMDM is first evaluated on unconditioned natural language generation using an MDLM backbone pretrained on OpenWebText \cite{Gokaslan2019OpenWeb} for 262B tokens \cite{sahoo2024simpleeffectivemaskeddiffusion}. Starting from the same pretrained backbone, the vanilla MDLM baseline is compared against SCMDM with both additive and concatenation-based fusion. All post-training runs use only 3.25B additional tokens, which is small relative to the original pretraining budget. For fair comparison, the reported vanilla MDLM results also include this additional post-training on the same data.

Language generation quality is evaluated from three complementary perspectives. First, generative perplexity is reported for unconditioned samples using two external autoregressive evaluators, GPT2-Large \cite{Radford2019LanguageMA} and Llama 3.2-1B \cite{grattafiori2024llama3herdmodels}. As shown in Table~\ref{tab:genppl}, SCMDM substantially improves unconditioned generation quality across most denoising steps under both evaluators and outperforms LDDM-M \cite{jo2026loopholingdiscretediffusiondeterministic}. The strongest gains come from the concatenation variant, which reduces GPT2-evaluated generative perplexity from 42.89 to 23.72 at 1000 denoising steps, with similarly large improvements at 128, 256, and 512 steps. The same trend holds under the Llama evaluator, where generative perplexity decreases from 46.90 to 24.96 at 1000 steps. We additionally evaluate a parameter-matched MDLM baseline,
$\text{MDLM}_{\text{param-match}}$, and find that SCMDM outperforms this baseline, demonstrating that its improvements are not a result of increased parameter count.
 Entropy and autoregressiveness statistics are additionally reported in Table~\ref{tab:entropy_ar_pretty}. These remain largely unchanged, indicating that the perplexity improvements are not explained by increased repetition or more autoregressive-like decoding behavior. Finally, Table~\ref{tab:zeroshot_ppl} shows that self-conditioned models also consistently improve zero-shot perplexity across standard downstream corpora and Table~\ref{tab:owt_ppl_262b} shows improved test perplexity for self-conditioned models on the OWT test dataset.

\begin{wraptable}[11]{r}{0.35\textwidth}
  \vspace{-1.5\baselineskip}
  \centering
  \setlength{\tabcolsep}{8pt}
  \renewcommand{\arraystretch}{1.05}
  \footnotesize
  \begin{tabular}{@{}l r@{}}
    \toprule
     & \textbf{PPL ($\downarrow$)} \\
    \midrule
    AR$^\dagger$        & 17.54 \\
    \midrule
    SEDD$^\dagger$      & $\leq$24.10 \\
    MDLM                & $\leq$23.73 \\
    \rowcolor{gray!15} SCMDM$_{\text{Add}}$    & $\leq$23.11 \\
    \rowcolor{gray!15} SCMDM$_{\text{Concat}}$ & \textbf{$\leq$23.10} \\
    \bottomrule
  \end{tabular}
   \caption{Test perplexities on OWT for models trained for 262B tokens. $\dagger$ denotes results from 
   \citeauthor{sahoo2024simpleeffectivemaskeddiffusion}.}
  \label{tab:owt_ppl_262b}
  \vspace{-0.6\baselineskip}
\end{wraptable}

To complement these distribution-based metrics, generated samples are also evaluated using an LLM-as-a-judge setup. Specifically, Gemma 31B \cite{manik2026gemma4phi4qwen3} is used to assign a single score from 0 to 100 based on overall text quality and coherence. The score distribution in Figure~\ref{fig:llm_judge} shifts to the right under SCMDM, with the mean increasing from 33.1 to 36.5, indicating improved generation quality. Additional details, including the judge prompt and qualitative samples, are provided in Appendix~\ref{app:llm_judge}.

A key result is that full self-conditioning is consistently more effective than partial self-conditioning in the post-training regime. While the $\text{rate}=0.5$ variant improves over vanilla MDLM, it remains substantially weaker than full self-conditioning across denoising budgets and evaluators. For example, under the GPT2 evaluator at 1000 steps, generative perplexity improves from 42.89 to 37.04 with $\text{rate}=0.5$, but to 23.72 with full self-conditioning. 
This supports the paper's central claim that, once the base denoiser is already informative, dedicating capacity to unconditional updates becomes detrimental.
Overall, these results show that SCMDM is an effective post-training adaptation for masked diffusion language models: it substantially improves unconditioned generation quality, zero-shot and test perplexity, and does so without increased autoregressive behavior, reduced token-level entropy, or major architectural complexity.

\begin{table}[t]
  \centering
  \setlength{\tabcolsep}{6pt}        
  \renewcommand{\arraystretch}{1.18} 
  \footnotesize
  \begin{tabular}{@{}l*{6}{>{\centering\arraybackslash}p{1.55cm}}@{}}
    \toprule
    & PTB & Wikitext & Lambada & AG News & Pubmed & Arxiv \\
    \midrule
    MDLM & 101.71 & 37.82 & 50.04 & 70.60 & 44.45 & 39.14 \\
    \midrule
    \rowcolor{gray!15}
    $\text{SCMDM}_\textrm{Add}$    & \textbf{98.59} & \textbf{36.46} & 49.10 & \textbf{68.03} & \textbf{41.72} & \textbf{37.67} \\
    \rowcolor{gray!15}
    $\text{SCMDM}_\textrm{Concat}$ & 99.13 & 36.48 & \textbf{49.09} & 68.13 & \textbf{41.72} & 37.75 \\
    \bottomrule
  \end{tabular}
  \caption{Zero-shot perplexities ($\downarrow$) of models trained for 262B tokens on OWT. MDLM and SCMDM were further finetuned another 3.25 billion tokens on OWT. All perplexities for diffusion models are upper bounds.}
  \label{tab:zeroshot_ppl}
\end{table}

\subsection{Biological Sequence Generation}
\label{subsec:exp_bio}

\textbf{Settings.} For the biological sequence experiments, two discrete domains are examined: genomes and small molecules. The reference genomes from ten-species dataset (Species10) were downloaded from
NCBI refseq database \cite{o2016reference}. A bidirectional Mamba model (4.8M parameters) was then trained on this dataset, similar to \cite{schiff2025simpleguidancemechanismsdiscrete}. The total number of training tokens is approximately 62B, while the post-training budget is one quarter of that.
For small molecules, the QM9 dataset is used \cite{ruddigkeit2012enumeration, ramakrishnan2014quantum}, where molecules are represented by SMILES strings \cite{weininger1988smiles}. The base model for this dataset has Transformer architecture with 92.4M parameters. The number of denoising steps (or number of function evaluations - NFE) for the molecule task is set at 32 as \cite{schiff2025simpleguidancemechanismsdiscrete}. With these two tasks, the concatenation-based fusion SCMDM is used for post-training the base models. The total number of training tokens is approximately 819M. A budget of 262M is used to post-train the model. The size of the standard post-trained models are scaled to be at least larger compared to SCMDM. Concatenation-based fusion is used for SCMDM for these tasks.

\begin{wraptable}[15]{r}{0.45\textwidth}
\centering
\vspace{-10pt}
\footnotesize
\setlength{\tabcolsep}{3pt}
\begin{tabular}{lcccc}
\toprule
& \multicolumn{2}{c}{NFE = 128 $[\downarrow]$} 
& \multicolumn{2}{c}{NFE=256 $[\downarrow]$} \\
\cmidrule(lr){2-3} \cmidrule(lr){4-5}
Method 
& 3-mer & 6-mer 
& 3-mer & 6-mer \\
\midrule
MDM                  & 1.77 & 4.71 & 1.85 & 4.89 \\
$\text{MDM}_{\text{param-match}}$    & 2.58 & 5.38 & 2.49 & 5.65 \\
$\text{SCMDM}_{\lambda=0.25}$    & 2.94 & 5.65 & 2.85 & 5.63 \\
$\text{SCMDM}_{\lambda=0.5}$     & 2.27 & 5.05 & 2.29 & 5.14 \\
$\text{SCMDM}_{\lambda=0.75}$    & 1.63 & 4.98 & 1.97 & 5.75 \\
\rowcolor{gray!15}
SCMDM                & \textbf{1.58} & \textbf{4.52} 
                     & \textbf{1.67} & \textbf{4.63} \\
\bottomrule
\end{tabular}
\vspace{4pt}
\captionof{table}{Genomic distribution fidelity of different methods on Species10. JS divergence for 3-mer and 6-mer distributions is reported, scaled by $10^{-2}$. Lower is better.}
\label{tab:genomic_self_conditioning_ablation}
\end{wraptable}

\textbf{Genome generation.} 
On the genome generation task (\Cref{tab:genomic_self_conditioning_ablation}), SCMDM with full self-conditioning consistently achieves superior distributional alignment compared to both the vanilla MDM and the post-trained MDM baseline across all denoising budgets. Specifically, at 128 steps, SCMDM yields the best performance with JS 3-mer and 6-mer scores of 1.58 and 4.52, significantly outperforming the standard model and the param-match baseline. These results indicate that the self-conditioning mechanism effectively enables the model to refine its predictions for complex biological motifs that standard MDMs struggle to capture. 
The table also isolates the effect of the self-conditioning against partial self-conditioning: it shows that the inclusion of unconditioned updates during post-training can be detrimental to performance, leading to higher JS divergence scores across most step budgets. These results support the post-training strategy: once the self-conditioning signal is informative, retaining it throughout finetuning is more effective than stochastically dropping it.

\begin{wraptable}[10]{r}{0.45\textwidth}
\centering
  \vspace{-10pt}
  \footnotesize
  \setlength{\tabcolsep}{5pt}
  \renewcommand{\arraystretch}{1.18}
  \begin{tabular}{@{}l r r@{}}
    \toprule
    & Valid & Novel \\
    \midrule
    
    $\text{MDM}$   & $594.2\pm9.5$ & $207.0\pm10.0$ \\
    $\text{MDM}_{\text{param-match}}$  & $582.8\pm19.5$  & $222.4\pm14.9$ \\
    $\text{SCMDM}_{\lambda=0.5}$ & $589.0\pm12.2$  & \bm{$225.6\pm13.8$} \\
    \rowcolor{gray!15}
    $\text{SCMDM}$ & \bm{$618.0\pm13.2$} & $180.4\pm14.6$ \\
    \bottomrule
  \end{tabular}
  \caption{SCMDM vs. baselines evaluated over 1024 generated molecules.}
  \label{tab:qm9_no_guidance}
\end{wraptable}

\textbf{Small molecule generation.}
For molecular generation on QM9, the number of denoising steps is fixed at 32 for all methods. As shown in \Cref{tab:qm9_no_guidance}, SCMDM with full self-conditioning achieves the highest validity with $618.0 \pm 13.2$ valid molecules, outperforming the base model ($594.2 \pm 9.5$) and the param-match baseline ($582.8 \pm 19.5$). However, a distinct performance trade-off is observed regarding novelty: while SCMDM excels in fidelity, its novelty score ($180.4 \pm 14.6$) is lower than other baselines. 
This suggests that for tasks with relatively compact search spaces like QM9, SCMDM maintains high validity through iterative refinement but becomes overly conservative, ultimately trading off novelty for fidelity. For other tasks that have much larger search spaces, novelty is not a problem because the number of possible combinations is extremely high. For example, in the genomic experiments, all runs of SCMDM generate 100\% non-duplicated and novel sequences. Meanwhile, the improved JS scores show that SCMDM maintains strong distributional alignment, meaning the generated samples better match the real data distribution.

\subsection{Discretized Image Generation}

\textbf{Settings.} For the image generation experiments, the models were trained on CIFAR-10 \cite{krizhevsky2009learning} using approximately 44B training tokens. The base model for this dataset has U-net \cite{ronneberger2015u} architecture with 36M parameters. The concatenation-based fusion SCMDM is used for post-training the base models. For post-training, roughly 20B training tokens are used. A total of 128 denoising steps is used for each method to calculate fréchet Inception Distance (FID) \cite{heusel2017gans}. The size of the standard post-trained models are scaled to be at least larger compared to SCMDM. Concatenation-based fusion is used for SCMDM for this task.

\begin{wraptable}[11]{r}{0.32\textwidth}
  \vspace{-1.9\baselineskip}
  \centering
  \footnotesize
  \setlength{\tabcolsep}{5pt}
  \renewcommand{\arraystretch}{1.05}
  \footnotesize
  \begin{tabular}{@{}l r r@{}}
    \toprule
     & NFE & \textbf{FID ($\downarrow$)} \\
    \midrule
    $\text{MDM}_{\text{Base model}}$     &  128 & 86.48 \\
    & 1000 & 40.57 \\
    \midrule
    $\text{MDM}_{\text{param-match}}$ & 128 & 82.02 \\
    & 1000 & 37.98 \\
    \midrule
    $\text{SCMDM}_\textrm{$\lambda$=0.5}$   &  128 & 80.22 \\
    & 1000 & 37.83 \\
    \midrule
    \rowcolor{gray!15} 
    $\text{SCMDM}$   &  128  & 78.59 \\
    \rowcolor{gray!15} 
    & 1000 & \textbf{37.81} \\
    \bottomrule
  \end{tabular}
   \caption{CIFAR-10 FID of SCMDM vs. MDM.}
     \label{tab:cifar_10}
  \vspace{-0.6\baselineskip}
\end{wraptable}

\textbf{Results.}
On CIFAR-10 image generation, SCMDM consistently improves over the standard MDM baseline in terms of FID across different self-conditioning configurations. As shown in \Cref{tab:cifar_10}, with NFE=128, SCMDM with $\lambda=0.5$ already yields a noticeable improvement, reducing FID from 86.48 to 80.22.
When fully removing unconditioning during post-training, SCMDM achieves the best performance with an FID of 78.59 (NFE=128) and 37.81 (NFE=1000), demonstrating that pure self-conditioning is more effective for improving sample quality in this setting. Moreover, both variants of SCMDM outperform the standard MDM of the same model size.

\section{Discussion}
\label{sec:discussion}

Across all four domains, full self-conditioning consistently outperforms partial dropout
($\lambda \in \{0.25,0.5,0.75\}$) when adapting pretrained MDMs. This contrasts with
self-conditioned diffusion models trained from scratch, where partial dropout is commonly used~\cite{chen2023analogbitsgeneratingdiscrete}.
The difference arises because post-training and training from scratch optimize different
problems. From scratch, both unconditional denoising and self-conditioned refinement are
initially unsolved, so mixing the two objectives can stabilize learning. In post-training,
however, the pretrained backbone has already learned the unconditional masked denoising
task. As a result, unconditional updates provide limited new signal, while self-conditioning
introduces a new refinement channel that the model must learn to exploit. Allocating
updates to the unconditional path therefore trades away refinement capacity.

The experiments support this interpretation. The largest gap appears for the strongest
pretrained backbone, MDLM-OWT trained for 262B tokens (Table~\ref{tab:genppl}). At
1000 denoising steps, GPT2-Large generative perplexity improves from $42.89$ for
vanilla MDLM to $37.04$ with $\lambda=0.5$, but reaches $23.72$ with full
self-conditioning. Thus, the gain from partial to full self-conditioning
($37.04 \to 23.72$) is more than twice the gain from vanilla to partial
self-conditioning ($42.89 \to 37.04$). Similar trends appear in the genomic and
CIFAR-10 experiments (Tables~\ref{tab:genomic_self_conditioning_ablation}
and~\ref{tab:cifar_10}). These results suggest that the commonly used $\lambda = 0.5$
heuristic is not well suited to post-training converged MDMs, and that the benefit of
full self-conditioning grows as the pretrained backbone becomes more capable.

\section{Conclusion}
\label{sec:conclusion}
This paper introduces \textit{Self-Conditioned Masked Diffusion Models} (SCMDM), a lightweight post-training adaptation for masked diffusion models that conditions each denoising step on the model’s own previous clean-state predictions. Across natural language, molecular generation, and genomic sequence modeling, SCMDM consistently improves over vanilla masked diffusion baselines with only minimal architectural change. We further show that self-conditioning behaves differently in the post-training regime: once the base denoiser becomes sufficiently informative, full self-conditioning is more effective than partial self-conditioning. Overall, these results establish simple clean-state self-conditioning as an effective and broadly applicable post-training strategy for improving pretrained masked diffusion models.

\bibliographystyle{unsrtnat}
\bibliography{example_paper}

\newpage
\appendix
\onecolumn
\section{Experimental Details}
\label{sec:exp_details}
\subsection{Natural Language}

\textbf{Entropy.}
We measure generation diversity by token-level entropy as defined in \cite{zheng2025maskeddiffusionmodelssecretly}. For a generated sequence of length $L$ with $K$ distinct tokens, let $L_k$ denote the count of token $k$, and define the empirical token probability as $p_k = L_k / L$. The entropy of the sequence is then
\[
H(\bx) = - \sum_{k=1}^{K} p_k \log p_k.
\]
Higher entropy means a more uniform token distribution and greater lexical diversity, whereas lower entropy indicates that probability mass is concentrated on fewer tokens, reflecting reduced diversity and potentially more repetitive samples.

\textbf{Autoregressiveness (AR-ness).}
To characterize how closely diffusion decoding resembles left-to-right autoregressive generation, we report both local and global autoregressiveness metrics as described in DiffuCoder~\cite{gong2025diffucoderunderstandingimprovingmasked}. Local AR-ness measures the tendency of the model to decode immediate successor positions in contiguous next-token order. Intuitively, Local AR-ness@1 captures how often the newly unmasked position directly continues a previously generated position, and higher values indicate stronger local left-to-right behavior. Global AR-ness measures the tendency to select one of the earliest remaining masked positions at each decoding step. In our experiments, we report Global AR-ness@4, which measures how often the newly unmasked position lies among the first four still-masked positions. Higher values therefore indicate a stronger preference for front-to-back filling and more autoregressive-like generation overall.

\textbf{LLM-Judge.}
\label{app:llm_judge}
For unconditioned natural language generation evaluation, in addition to the standard distribution matching metrics, we evaluate with LLM-as-a-Judge. The objective is to quantify the \textit{overall text quality} and \textit{coherence} of the generated text. This method is commonly used, and has been shown to generally match human evaluation performance \cite{zheng2023judgingllmasajudgemtbenchchatbot}. For the results shown in Section \ref{exp:NL}, the Judge LLM used is Gemma 4 with 31 Billion parameters \cite{manik2026gemma4phi4qwen3}.

\textbf{Judge Prompt}. We use the following prompt to score unconditioned text generations for overall text quality and coherence.

\begin{lstlisting}[style=promptstyle]
You are evaluating an unconditioned text generation sample from a language model.

Rate the sample with one overall score from 0 to 100 based only on overall text quality and coherence.

A high score means the text is clear, fluent, and coherent.
A low score means the text is disjointed, confusing, repetitive, or nonsensical.

Return only valid JSON in exactly this format:
{
  "overall_score": <integer 0-100>
}

Text sample:
---
{sample_text}
---
\end{lstlisting}

\subsection{Biological Sequence and Image Generation}

We evaluate our method on biological sequence generation tasks using the Species10 and QM9 datasets; and image generation task with CIFAR-10 datasets. The detailed training and post-training hyperparameters are summarized in Table~\ref{tab:non_language_setting}. All experiments were conducted on NVIDIA A6000 GPUs. 

For Species10, we train the model for 120K steps with a large context size of 32,768 and a batch size of 32, reflecting the long-range dependencies in genomic sequences. In contrast, QM9 involves shorter sequences, and we adopt a smaller context size of 32 with a larger batch size of 2048, training for 25K steps. For CIFAR-10, we train the model for 600K steps with a context size of 3072 (32 $\times$ 32 $\times$ 3).

We use a learning rate of $2e^{-3}$ for Species10, $3e^{-4}$ for QM9, and $2e^{-4}$ for CIFAR-10 during training, with linear warmup over 3K, 1K, and 5K steps, respectively. In the post-training stage, we adopt a unified learning rate of $3e^{-4}$ for both datasets. All models are optimized using ADAM with $(\beta_1, \beta_2) = (0.9, 0.999)$, similar to \cite{schiff2025simpleguidancemechanismsdiscrete}.

\begin{wraptable}[14]{r}{0.5\textwidth}
    \centering
    \footnotesize
    \caption{Training and post-training hyperparameters for biological sequence and image generation experiments.}
    \renewcommand{\arraystretch}{1.25}
    \begin{tabular}{|c|c|c|c|}\hline
         Datasets&  Species10& QM9 & CIFAR-10\\\hline
         Train
steps &  120K& 25K& 600K\\\hline
         Context size&  32768& 32& 3072\\\hline
         Batch size&  32& 2048 &512 \\\hline
         Training LR&  $2e^{-3}$& $3e^{-4}$&$2e^{-4}$\\\hline
         LR warmup
steps &  3K& 1K& 5K\\ \hline
 Post-training LR& \multicolumn{3}{c|}{$3e^{-4}$} \\\hline
         Optimizer&  \multicolumn{3}{c|}{ADAM (0.9, 0.999)}\\\hline
         
    \end{tabular}
    \label{tab:non_language_setting}
\end{wraptable}

For the genomics setting (Species10), we evaluate generation quality using \textbf{Jensen--Shannon (JS)} divergence over $k$-mer distributions, specifically 3-mer and 6-mer statistics. We generate 1000 samples to measure these evaluation metrics. These metrics capture sequence-level patterns at different granularities: 3-mers reflect local nucleotide composition, while 6-mers encode longer-range dependencies and motif-level structure, providing a more comprehensive assessment of biological realism.

For QM9, we adopt standard molecular generation metrics. \textbf{Validity} count the number of generated molecules that correspond to chemically valid structures. \textbf{Uniqueness} is computed as the fraction of distinct molecules among the valid set (i.e., removing duplicates within valid generations). \textbf{Novelty} is defined as the proportion of unique molecules that do not appear in the training dataset, reflecting the model’s ability to generate previously unseen compounds. We run each experiments 5 times, each time generating 1024 samples, similar to \cite{schiff2025simpleguidancemechanismsdiscrete}.

For image generation, we use \textbf{Fréchet Inception Distance (FID)} to compare results, which is the standard and widely adopted measure for assessing sample quality in image generation tasks, providing a reliable estimate of both fidelity and diversity \cite{cao2024survey}. We generate 50,000 images for each measurement.

\section{Additional Results}

\subsection{Qualitative Samples}
For qualitative inspection, we present unconditional generations from the vanilla MDLM baseline and $\text{SCMDM}_\text{Concat}$ under matched sampling settings and fixed random seeds. For brevity, we show excerpted portions of the generations rather than the full sequences. For these samples we further score them with GPT 5.4 as the LLM judge, using the same prompt as written in Appendix \ref{app:llm_judge}. For the scoring, the entire sample output is used, rather than the excerpt shown. Notably, for both included samples, the $\text{SCMDM}_\text{Concat}$ model achieves a higher score for quality/coherence when compared to the vanilla MDLM baseline.

\newpage
\noindent\textbf{Sample 1 (Seed 9). }

\noindent\makebox[\linewidth][c]{%
\begin{minipage}[t]{\linewidth}
\vspace{0pt}%
\begin{minipage}[t]{0.49\linewidth}
\vspace{0pt}%
\begin{samplebox}[Vanilla MDLM (GPT 5.4 Rating: 18/100)]
<|endoftext|> a player at times to be is better member of the team. So I felt a spark off of him with that, and I think I put something short to Bo Ryan, I thought that was a great play because we called out or something about how somebody I think could take a hit, do a good job doing those physical things so that he could get something outside of what he knows are he has. So I made no mistake it was me, so I kind of want to be a little younger than Bo Ryan. I think you put some skill value in him so you level up the speed and he can, obviously, a good guy, an international baseball player, so I really like Sings. I think I understand his maturity because I’m Canadian, I think I know how I like a lot of Puerto Rican National League prospects and I think they were smart. I think there’s potential that that would play to his offense. You see his personality, the personality, definitely makes his love of the game obvious. I just enjoyed hearing stuff. I would have it arrived at his house at the Wahoo. The guys here own St. Louis …

You know, you know, are you the Red Sox

Wahoo, because you tend to play golf and sit and host their house with your family … Yeah, so yeah, I had I been with him no would have talked it up because ESPN is not putting him on their couch. In fact, they put him on their own ESPN screen So I had to see him in the house. Even before all that, I had been to him show up in Philadelphia with Matt, and he tells me he’s really excited about playing in the Orchard now, because if it were not for him, before I left his house on the Saints, it would be a little weird. But I think he would be stud guys right now.

How would you want to play football or again, or will the injury give you some time to return to the lineup, when the Rangers have performed well all along?

\end{samplebox}
\end{minipage}\hfill
\begin{minipage}[t]{0.49\linewidth}
\vspace{0pt}%
\begin{samplebox}[$\text{SCMDM}_\text{Concat}$ (GPT 5.4 Rating: 43/100)]
<|endoftext|> a player at the position be is the receiver of the team. So I have a little bit of experience with that, and I think that in the short term he could get better in both areas of the play because we has great depth with us, and he’s a good athlete and a good player who is physical, so definitely he could get a lot of what he knows in the backfield.

Who do you look up to as a running back? Will it be Metteneau or Ryan Tannehill?

Ryan is definitely the franchise quarterback that he is, and I think he’s going to be an excellent football player, so I feel like not only can I get to understand his personality but I’m going to like him in total. I think that one of the things I appreciate about Metteneau is he’s a high-energy athlete that has to play to his strengths. I see his attitude on the field. I see his love of sport. I see his passion and his game. I absolutely have great respect for his personality and his personality, so it’s a great way to get to understand it.

What do you think about the ESPN media coverage? Was it your idea to play golf and a break or just go with your family?

I think it’s been really interesting for him to do it just because ESPN is playing with him, supported by the very fact that he is on national television on ESPN Radio. I had experiences with him in the NFL, and because of that, I’ve been able to handle him a lot, and I also feel that it would be really fun actually playing in the NFL, and I think if you’re hearing up before he plays, they’re going to tune in a little bit. So I think he would be a really good athlete.

How would you want to play?

He’s got to get some runs, to finally get some, luckily the Patriots have a great running back on their bench and we also have good depth with us, so obviously, that will be a great way to get him going.

\end{samplebox}
\end{minipage}
\end{minipage}%
}

\vspace{0.8em}

\newpage
\noindent\textbf{Sample 2 (Seed 100).}

\noindent\makebox[\linewidth][c]{%
\begin{minipage}[t]{\linewidth}
\vspace{0pt}%
\begin{minipage}[t]{0.49\linewidth}
\vspace{0pt}%
\begin{samplebox}[Vanilla MDLM (GPT 5.4 Rating: 14/100)]
<|endoftext|> days ago, less than the power needed for water and electricity.

As of 15 February, one out of every six customers connected to a network were calling in 60 minutes, twice daily times. In total, 7.7 million cell cycles belonged to the phones data, or about 8 million.

The problem is most severe within the rural communities, which rise to 8$\%$ of households, and especially high in the regions (16$\%$) of further cities.

About 16$\%$ of households live in automobile regions, with all available transport services covered fully, which includes auto travel (13$\%$) and taxis (9$\%$-10$\%$).

Source: Reuters<|endoftext|>So what of a “too much alternative game”?

No!

Do you want to?

Who’s the game, yourself?

But that’s what political people in politics think.

And ultimately you’d better – to best – answer this question, as did Wes Bonchug, a media strategist known as shadow-runner Bernie Sanders.

“Too much people cashing in touching on you almost right out the way.”

One of the campaign’s most memorable moments Democratic presidential nominee Hillary Clinton — rather her father Cardinal Schlegel) — might give during a studio audience.

As part of a recent meeting, Clinton was invited to speak at a venue we’ll call a “charinal hall,” with her quoted in his January 14, 2016 speech.

A director of the documentary God’s Gazing, Sanders saw no place in a high-profile arena like that.

“There were a lot of candidates who had a reputation,” Sanders told Bonchug.

“I think the media does the job too well and you don’t get enough respect and you don’t want enough people with you. You can’t hire people.”

“How about people like you, who always thought they’d be there to help you,” he added.

“But we’ve got so many stories from one old New York video that’s photograph of you, by a pedestrian, one, by a woman. All the way of press tour. I mean let’s think about it.

JESSICA WARMACHINE, CHERATE AGE

\end{samplebox}
\end{minipage}\hfill
\begin{minipage}[t]{0.49\linewidth}
\vspace{0pt}%
\begin{samplebox}[$\text{SCMDM}_\text{Concat}$ (GPT 5.4 Rating: 52/100)]
<|endoftext|> we still believe in a Unix-based system, and we believe that the popularity of that system will continue to grow. Our customers want to support Linux that’s mature, and they always see the benefits of Linux.

One of the really important things about Linux in Ubuntu 8 and the statement that it communicates is, “Lately if you support a mature process, your product will become mature.” So it’s been that clear.

When I told my customers, I said, “I’ve already started working on Ubuntu Enterprise (Linux). It will be great for people.”

And they disagreed with what I said. And I think their argument about that point was everything that this market required; you had to sell a simple product, and once the game worked, you had to have that and not have bugs. So they didn’t have a week, even maybe 30 days, to release a release. Instead, you just had to sell an implementation that was really mature, by which I really meant.

“Pretty much every buyer fits in based on being made in the right way.”

One thing that I got also isn’t just a sales experience, but rather a commercial (openSUSE) experience. It, in itself, is a really critical part of the mission of the company. The difference to having a machine-level experience is so important because, how do you explain that? You look at a lot of things about Linux — its scale, its machine management stack — these are really intricate things, and the machine management stack structure has a structure, so being done in the right way is really hard.

MS: What do you struggle with most when it comes to an OS?

LG: I think that people understand the experience too well. And don’t get me wrong, usually I just add a way to interact with machines so they can better understand Linux. But there have been a lot of changes just in the last years, and I thought I’d give you a chance to explain things from a different perspective.

\end{samplebox}
\end{minipage}
\end{minipage}%
}

\subsection{Number of Model Parameters}

\begin{table}[t]
\centering
\footnotesize
\caption{Number of model parameters for each task.}
\label{tab:parameter_increase}
\begin{tabular}{lrrr}
\toprule
\textbf{Task} & \textbf{Base Model Params} & \textbf{Concat Model Params} & \textbf{Concat \% Increase} \\
\midrule
Language & 169,627,250 & 209,405,042 & 23.45\% \\
Genome (Mamba) & 4,774,912 & 4,909,056 & 2.81\% \\
Small molecule (Transformer) & 92,442,152 & 93,652,520 & 1.31\%\\
Discretized image (U-net) & 35,749,766 & 35,783,174 & 0.10\%\\
\bottomrule
\end{tabular}
\end{table}

\Cref{tab:parameter_increase} reports the parameter counts for the base MDM backbones and their SCMDM counterparts using concatenation-based fusion. Consistent with our goal of providing a lightweight adaptation, the architectural overhead introduced by SCMDM is minimal across most domains. For the genomic, molecular, and image generation tasks, the parameter increase remains below $3\%$, with the U-net backbone for image generation incurring a negligible overhead of only $0.10\%$. For natural language generation, the increase is larger at roughly $23\%$, because the self-conditioning module must project a large vocabulary-sized distribution into the model hidden space. For fair comparison, we also scale MDMs' model to be at least the same as the number of parameters as SCMDM:

\begin{itemize}

    \item \textbf{Natural Language:} We augment each DiT block with a supplementary residual MLP branch of the form $\text{MLP}(h) = W_2 \text{GELU}(W_1 h)$, where $W_1 \in \mathbb{R}^{H \times r}$ and $W_2 \in \mathbb{R}^{r \times H}$. Here, $H$ denotes the hidden dimension of the backbone, and $r$ is the bottleneck width chosen to match the parameter increase introduced by SCMDM. For the small MDLM used in our language experiments, we set $H=768$ and $r=2304$, which yields a total parameter count of approximately 212M.

    \item \textbf{Genome:} We integrate a lightweight residual MLP bottleneck within each block of the DiMamba backbone. This auxiliary branch is structured as $\text{MLP}(h) = W_2 \text{GELU}(W_1 h)$, where $W_1 \in \mathbb{R}^{H \times r}$ and $W_2 \in \mathbb{R}^{r \times H}$. For this example we set $r=34$. The total number of parameters in this scaled model is 4,916,496.
    \item \textbf{Small molecules:} We augment each DiT block with a supplementary residual MLP branch. We set the bottleneck dimension $r=66$. The total number of parameters in this scaled model is 93,668,672.
    \item \textbf{Discretized image:} We add a $1 \times 1$ residual bottleneck branch to each ResBlock. This branch follows the sequence $\text{Conv}_{1 \times 1}(c \to r) \to \text{GELU} \to \text{Conv}_{1 \times 1}(r \to c)$, where the channel dimension $c$ corresponds to the hidden dimension, and $r$ is the bottleneck dimension used to scale the parameter count. We set $r=3$.  The total number of parameters in this scaled model is 35,784,776. 
\end{itemize}

Notably, as shown in \Cref{sec:experiments}, these scaled models perform worse than SCMDM, though they have more parameters.

\section*{Broader Impacts}
\label{app:broader_impact}

This work improves post-training adaptation for masked diffusion models across language, image, molecular, and genomic domains. Potential benefits include more efficient reuse of pretrained models and improved generative modeling for scientific applications. Potential risks include misuse of stronger generative models for misleading synthetic content or sensitive biological and molecular design tasks. We do not release a deployed system or high-risk dataset, but encourage responsible use, domain-specific validation, and adherence to relevant safety and data-governance practices.

\section*{Limitations}

SCMDM is evaluated as a post-training adaptation for pretrained masked diffusion models, and its benefits may depend on the quality of the underlying denoiser. If the base model's clean-state predictions are weak, self-conditioning may provide less useful refinement signal. The method also introduces additional parameters, with a larger overhead for language models under concatenation-based fusion. While sampling requires no extra denoiser evaluations, post-training uses a two-pass update and therefore increases training cost. Finally, in molecular generation, SCMDM improves validity but reduces novelty on QM9, suggesting that self-conditioning can make generation more conservative in compact output spaces.

\end{document}